\documentclass{egpubl}
\usepackage{pg2020}
 
\usepackage[T1]{fontenc}
\usepackage{dfadobe}  

\biberVersion
\BibtexOrBiblatex
\usepackage[backend=biber,bibstyle=EG,citestyle=alphabetic,backref=true]{biblatex} 
\electronicVersion
\PrintedOrElectronic

\ifpdf \usepackage[pdftex]{graphicx} \pdfcompresslevel=9
\else \usepackage[dvips]{graphicx} \fi

\usepackage{egweblnk}

\usepackage{amssymb}
\usepackage{amsmath}
\usepackage{multirow}
\usepackage{xcolor}
\usepackage[left]{lineno}
\usepackage{soul}
\usepackage{float}
\usepackage{hyperref}
\usepackage{comment}

\graphicspath{{./images/}}
\title[Pixel-wise Dense Detector for Image Inpainting]%
      {Pixel-wise Dense Detector for Image Inpainting}

\author[R.Zhang et al.]
{\parbox{\textwidth}{\centering 
        Ruisong Zhang$^{1,2}$\orcid{0000-0002-7051-3309}, 
        Weize Quan$^{1,2}$\orcid{0000-0003-0892-581X}, 
        Baoyuan Wu$^{3,4}$\orcid{0000-0003-2183-5990}, 
        Zhifeng Li$^5$ %\orcid{0000-0001-7756-0901}
        and Dong-Ming Yan$^{1,2}$\thanks{Corresponding Author}\orcid{0000-0003-2209-2404}
        }
        \\
{\parbox{\textwidth}{\centering $^1$National Laboratory of Pattern Recognition, Institute of Automation, Chinese Academy of Sciences, Beijing 100190, China\\
         $^2$School of Artificial Intelligence, University of Chinese Academy of Sciences, Beijing 100049, China \\
         $^3$School of Data Science, the Chinese University of Hong Kong, Shenzhen, China\\
         $^4$Secure Computing Lab of Big Data, Shenzhen Research Institute of Big Data, China\\
         $^5$Tencent AI Lab, Shenzhen, China
       }
}
}
\begin{document}

\teaser{
    \centering
    \includegraphics[width=0.95\textwidth]{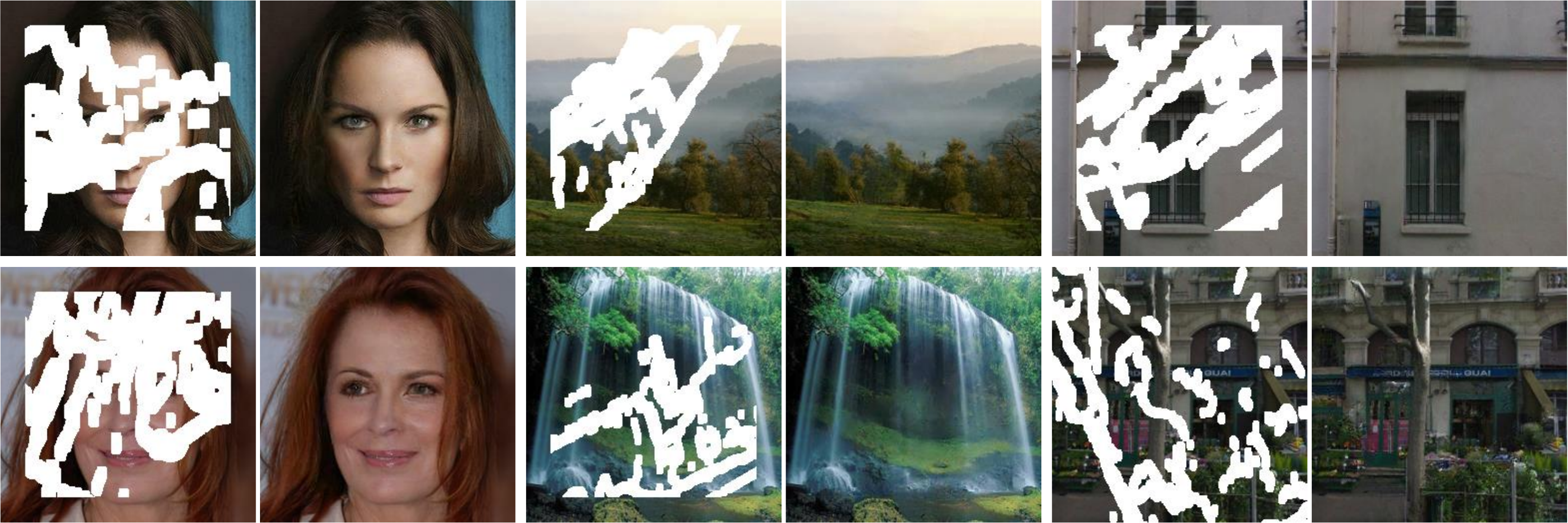}
    \caption{Image inpainting results by our proposed method.}
    \label{Fig:Intro}
}

\maketitle
%-------------------------------------------------------------------------
\begin{abstract}
   Recent GAN-based image inpainting approaches adopt an average strategy to discriminate the generated image and output a scalar, which inevitably lose the position information of visual artifacts. Moreover, the adversarial loss and reconstruction loss (\textit{e.g.}, $\ell_1$ loss) are combined with tradeoff weights, which are also difficult to tune. In this paper, we propose a novel detection-based generative framework for image inpainting, which adopts the min-max strategy in an adversarial process. The generator follows an encoder-decoder architecture to fill the missing regions, and the detector using weakly supervised learning localizes the position of artifacts in a pixel-wise manner. Such position information makes the generator pay attention to artifacts and further enhance them. More importantly, we explicitly insert the output of the detector into the reconstruction loss with a weighting criterion, which balances the weight of the adversarial loss and reconstruction loss automatically rather than manual operation. Experiments on multiple public datasets show the superior performance of the proposed framework. The source code is available at \url{https://github.com/Evergrow/GDN\_Inpainting}.
%-------------------------------------------------------------------------
%  ACM CCS 1998
%  (see https://www.acm.org/publications/computing-classification-system/1998)
% \begin{classification} % according to https://www.acm.org/publications/computing-classification-system/1998
% \CCScat{Computer Graphics}{I.3.3}{Picture/Image Generation}{Line and curve generation}
% \end{classification}
%-------------------------------------------------------------------------
%  ACM CCS 2012
%   (see https://www.acm.org/publications/class-2012)
%The tool at \url{http://dl.acm.org/ccs.cfm} can be used to generate
% CCS codes.
%Example:


\end{abstract}  
%-------------------------------------------------------------------------
\section{Introduction}

Image inpainting is a technique of filling the semantically correct and visually plausible contents in the missing regions of corrupted images, as shown in Fig.~\ref{Fig:Intro}. It has various applications, such as repairing deteriorated photographs, removing undesired objects from images, and even editing specified contents of images. Different from general generative tasks, image inpainting deals with corrupted images with plenty of contextual background, which is not only prior information to assist in reconstruction but also is a constraint to limit generated contents.
% The purpose of image inpainting is to reconstruct missing regions with visually realistic details and plausible structures.

Early works attempt to fill missing regions with some optimization algorithms, \textit{e.g.}, propagating the isophotes from boundaries~\cite{ballester2001filling,bertalmio2000image,efros2001image} or copying the matching information from background patches into missing regions~\cite{barnes2009patchmatch,darabi2012image,huang2014image}. These methods with low-level features achieve good results especially inpainting background or some repetitive patterns. However, as they cannot extract high-level semantic information, they often fail to generate reasonable structures with novel patterns in real-world scenarios. Moreover, high computational cost also limits their deployment in practical applications.

\begin{comment}
\begin{figure}[t]
  \centering
  \includegraphics[width=0.95\textwidth]{Intro1.pdf}
    \caption{Qualitative comparison of inpainting results by Partial Convolution (PConv)~\cite{liu2018image}, Pyramid-context Encoder Network (PEN)~\cite{zeng2019learning}, Gated Convolution (GConv)~\cite{yu2019free}, and Ours.}
  \label{Fig:Framework}
\end{figure}
\end{comment}

With the advance of deep learning, \textit{i.e.}, convolutional neural networks
(CNN) and generative adversarial networks (GAN)~\cite{goodfellow2014generative}, recent works~\cite{pathak2016context,yang2017high,iizuka2017globally,liu2018image,shiftnet2018,zeng2019learning} model image inpainting as a conditional generation problem to learn the mapping between the corrupted input images and the ground-truth images.
These deep inpainting techniques extract rich semantic information from large scale training data, based on contextual background, to fill missing regions with plausible contents and textures. Joint adversarial training also improves the visually realistic effect of image recovery. Unfortunately, these methods often suffer from distorted structures, blurry artifacts and distinct incoherence with surrounding areas, especially for complex scenes.

To address above problems, some deep inpainting methods~\cite{yu2018generative,yu2019free,nazeri2019edgeconnect,ren2019structureflow} utilize two-stage networks that rough out the missing structures or contents in the first stage, and then recover refined textures using coarse information in the second stage. Contextual attention~\cite{yu2018generative} inserts the attention module into the second stage to encourage spatial coherency of attention. Based on the structure image from the first stage, StructureFlow~\cite{ren2019structureflow} generates the textures to fill missing regions, which shows reasonable structures and vivid details. Compared with single encoder-decoder networks, however, two-stage networks are much deeper and thus cause extra computational cost and inference time.
%composite edge map
%Furthermore, partial convolution~\cite{liu2018image} is a masked convolution operation followed by a mask-update step designed for image inpainting task. It encourages to reduce artifacts disturbed from missing pixels. However, the non-differential property of updated mask makes the end-to-end training more difficult.

In addition, most deep inpainting methods~\cite{pathak2016context,iizuka2017globally,yu2018generative} follow an adversarial framework, which not only creates realistic results but also provides an agonizing multi-objective optimization. The first objective is the coherence and similarity with ground-truth images formulated as the reconstruction loss, and another objective is the perceptually realistic results programmed as the adversarial loss. Several tradeoff parameters are empirically set up to optimize these two objectives simultaneously. To our best knowledge, there are seldom works that improve this strategy with explicable theories.

To address the limitation discussed above, we propose a novel inpainting framework, comprising a generative model and a detective model, with a unique objective function to optimize.
Generative network is an encoder-decoder architecture with residual blocks~\cite{he2016deep} for repairing the corrupted images. Inspired by the image segmentation task~\cite{long2015fully}, we design and implement a fully convolutional network as detective network to evaluate the inpainting results in a pixel-wise manner.
The binary mask is the target of the detective network using an approach of weakly supervised learning to capture visual artifacts of entire generated image.
Compared with a single scalar from the standard discriminator, the location information of artifacts with up-sampling builds a dense mapping function between the output image and ground-truth image for more accurate valuation. Under the guidance of accurate valuation, inpainting techniques pay more attention to artifacts such as distortion, blurriness and incoherence, and reduce them, thus making the valuation from the detector trend inaccurate. The adversarial learning of the generator and detector moves from ``whether'' to ``where''. Moreover, we propose a reconstruction loss weighted the valuation as inpainting objective function to solve the multi-objective optimization. This loss function without the hyper-parameter is better for describing the image inpainting task. The optimal balance between vraisemblance  and similarity is taken by networks rather than empiricism. Our contributions are summarized as follows: 

%The optimization involving balanced loss removes the influence of different size of the mask regions at the same time.

%Experiments compared with state-of-the-art methods on standard datasets including faces (CelebaHQ~\cite{liu2015deep,karras2017progressive}) and natural images (Places2~\cite{zhou2017places}) demonstrate that the proposed method achieves competitive results. Furthermore, we conduct ablation studies to verify our modifications.

\begin{itemize}
    \item We propose a novel framework that merges the generator and the detector, and both experimental results and explicable theory show that this framework contributes to reduce artifacts.
    %\item We propose a novel framework merged the generator and the detector to reduce artifacts and inprove inpainting results.
    \item The detector using weakly supervised learning makes a dense estimate of the entire inpainted image, which is similar to human perceptual evaluation.
    %\item The detector makes a precise estimate of the entire generated image. The input without the ground-truth image is suitable for valuation during the test period, especially the case without pair original images.
    \item We design a weighted reconstruction loss to optimize the combined inpainting objectives including vraisemblance and similarity automatically, which achieves superior results.
    %\item Proposed optimization to the generator with weighted reconstruction loss automatically sovle combined problems of vraisemblance and similarity in image inpainting simultaneously.
\end{itemize}

% \item We train the detector using balanced loss to eliminate the interference from different missing regions in corrupted image.

\section{Related Work}

Existing image inpainting approaches can be classified into two categories: low-level features based approaches and deep semantic features based approaches. The former usually involves some geometric techniques for texture synthesis or structure propagation in low-level features. The latter often solves the inpainting problem by deep neural networks to extract global semantic features.

\subsection{Methods Based on Low-level Features}
Approaches based on low-level features are roughly divided into diffusion-based methods and patch-based methods.
Traditional diffusion-based methods ~\cite{ballester2001filling,bertalmio2000image,efros2001image} typically utilize variational algorithms to propagate neighboring appearance information (\textit{e.g.}, the isophotes) into the missing regions. Due to the limited extended prediction of partial differential equation, these methods could not produce good results when the missed regions are broad. The restoration regions generated by this kind of methods also lack meaningful structure information.

Unlike diffusion-based methods just focusing on the surrounding pixels of missing regions, patch-based methods~\cite{darabi2012image,huang2014image,danon2019unsupervised} measure the similarity between missing regions and each patch from the whole context, and recover target region by copying the matched patch. Bidirectional similarity measure~\cite{simakov2008summarizing} is proposed to model visual data for summarization and reduce visual artifacts. However, dense computation of patch similarity often comes at an expensive computation cost. To accelerate these searching algorithms, PatchMatch~\cite{barnes2009patchmatch} captures patch matches via random sampling and natural coherence in the imagery, and it is widely used in the interactive editing tools. Low-level features based approaches lacking deep understanding of whole image just generate repetitive content from background without unique filling information.

\subsection{Methods Based on Deep Semantic Features}
Deep semantic features based approaches attempt to perceive the semantic structure of the corrupted image by deep neural networks for better restoration results. Context Encoders~\cite{pathak2016context} first introduce CNNs for inpainting missing regions. The proposed encoder-decoder architecture is trained via incorporated reconstruction loss and adversarial
loss~\cite{goodfellow2014generative}. However, this network excessively concerns about entire consistency and it often results in visual artifacts in detailed regions. To generate high-frequency details, Yang \textit{et al.}~\cite{yang2017high} propose a multi-scale neural patch synthesis based on joint optimization of image content and texture constraints, and Iizuka \textit{et al.}~\cite{iizuka2017globally} unite global and local discriminator to assess completed image from generative network with dilated convolutions~\cite{yu2015multi}. However, local discriminator fails to deal with irregular missing regions.
%\revise{Moreover, this method requires additional post-processing to eliminate the visual inconsistency in the results output from the network.}
%, which is the baseline for most feature learning inpainting

Recently, deep inpainting techniques show the multiplex development. Attention is an important mechanism for image inpainting to build long-term correlations between missing regions and distant contextual information~\cite{yu2018generative,liu2019coherent,zeng2019learning,yu2019free}. Yu \textit{et al.}~\cite{yu2018generative} design a coarse-to-fine network and first introduce contextual attention into refined network. However, effect of attention mechanism mainly depends on results of coarse network, and poor coarse reconstruction often causes wrong match. To avoid the interference of corrupted regions, some works modify conventional convolutional operation, such as partial convolution~\cite{liu2018image} and gated convolution~\cite{yu2019free}, calculating convolution only on valid pixels. These variants succeed on reduction of blurry artifacts. EdgeConnect~\cite{nazeri2019edgeconnect} and StructureFlow~\cite{ren2019structureflow} generate reasonable structures with prior information and then synthesize fine texture. This easy-to-difficult process can obtain satisfactory visual effects, but it also needs sophisticated preprocessing to extract edge maps~\cite{nazeri2019edgeconnect} or edge-preserved
smooth images~\cite{ren2019structureflow}. In addition, several works~\cite{liu2018image,nazeri2019edgeconnect,xie2019image} project images into high-dimensional features space built by pretrained VGG-16~\cite{simonyan2014very} on ImageNet~\cite{deng2009imagenet} and then measure the similarity by perceptual loss~\cite{johnson2016perceptual} and style loss~\cite{Gatys_2016_image} to improve inpainting results. A possible limitation is that it reduces filling generalization of the scene outside ImageNet~\cite{deng2009imagenet}.

Deep inpainting approaches reviewed above mostly follow the adversarial framework used in Context Encoders~\cite{pathak2016context}. In this framework, the discriminator takes the inpainted image as the input to evaluate in level of whole image or its patches (\textit{e.g.}, PatchGAN~\cite{isola2017image}) , and discards the meaningful location information of blurry artifacts in the adversarial loss when training the generator. Specially, in PatchGAN, the patch-level valuations constitute a tensor, which is unseen for the generator, while the generator can only access the adversarial loss after average. On the contrary, we introduce a detective network to detect artifacts pixel by pixel, and the output of the detector assists the generator in eliminating color discrepancy and blurriness via an adaptive weighting strategy.

%from inpainting localization problem~\cite{li2019localization}

\section{Proposed Method}
\label{Sec:method}

\begin{figure*}[t]
  \centering
  \includegraphics[width=0.94\textwidth]{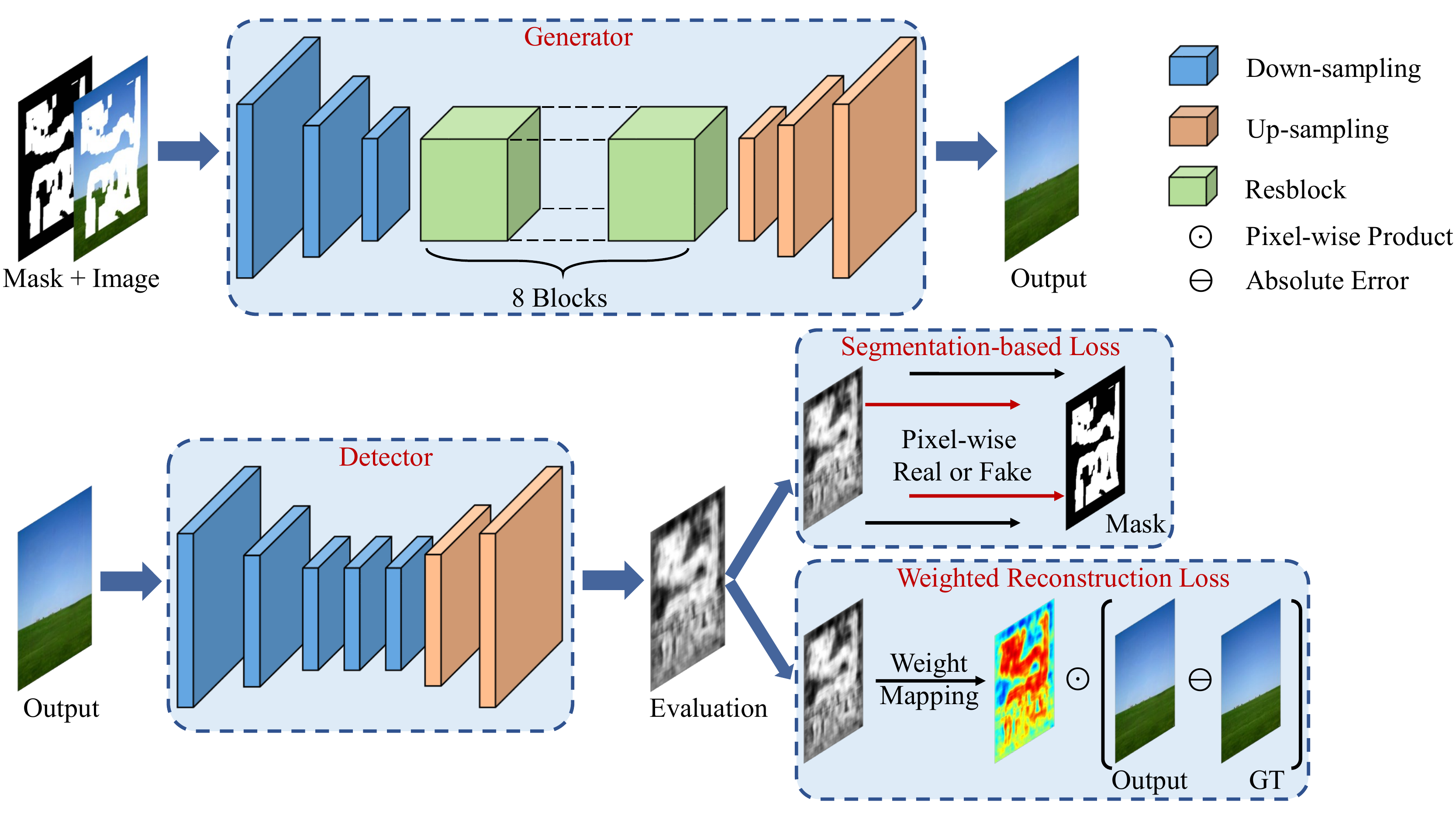}
  \caption{Illustration of our proposed detection-based generative framework.}
  \label{Fig:Framework}
\end{figure*}

Our proposed method utilizes a generative network to reconstruct corrupted images and a detective network to evaluate outputs of the generator to perform image inpainting, as shown in Fig.~\ref{Fig:Framework}. In this section, we first review GAN-based image inpainting. Then, we present the proposed novel detection-based generative framework, \textit{i.e.}, the coupling of the generator and the detector in training stage. At last, the details of network architecture and the loss function of our method are explained.

\subsection{GAN-based Image Inpainting}
\label{sec:review}

Generative Adversarial Network~\cite{goodfellow2014generative} is an advanced generative framework including two nets: generator $G$ and discriminator $D$. The contest training strategy drives both networks to improve their performance until reaching to a global optimum. Mathematically, this competitive process between $G$ and $D$ can be described as a min-max optimization problem with value function $V(G, D)$:
\begin{equation}
    \label{Eq.GAN}
    \begin{split}
    \min_G \max_{D} V(G, D) &= \mathbb{E}_{\mathbf{x} \sim p_{data}(\mathbf{x})} [\log D(\mathbf{x})] \\
    &+ \mathbb{E}_{\mathbf{z} \sim p_{\mathbf{z}}(\mathbf{z})} [\log(1 - D(G(\mathbf{z})))],
    \end{split}
\end{equation}
where $\mathbf{x}$ is real data, and $\mathbf{z}$ is input noise. The distribution is omitted for simplicity in following formulas. In Eq.~\eqref{Eq.GAN}, the discriminator $D$ tells the difference between real data distribution $\mathbf{x}$ and generated data distribution $G(\mathbf{z})$ as much as possible. At the same time, the generator $G$ tries to mimic real data distribution and fool the discriminator $D$.

Image inpainting is a special generative problem with plenty of prior information (\textit{i.e.}, corrupted image) as input rather than random noise. To improve the reconstruction quality of corrupted regions, and make whole image more realistic and vivid, most of deep image inpainting methods follow the GAN-based framework. The corresponding optimization for image inpainting is written as:
\begin{equation}
    \label{Eq.adv_loss}
    \min_G \max_{D} V(G, D) = \mathbb{E} [\log D(\boldsymbol{\mathrm{I}}_{gt})] + \mathbb{E} [\log(1 - D(G(\boldsymbol{\mathrm{I}}_{in}, \boldsymbol{\mathrm{M}})))],
\end{equation}
where $\boldsymbol{\mathrm{I}}_{gt}$ is the ground-truth image, $\boldsymbol{\mathrm{I}}_{in}$ is the input corrupted image, and $\boldsymbol{\mathrm{M}}$ is the binary mask (1: the missing region and 0: valid regions). Usually, $\boldsymbol{\mathrm{I}}_{in} = \boldsymbol{\mathrm{I}}_{gt} \odot (1 - \boldsymbol{\mathrm{M}}) + \boldsymbol{\mathrm{M}}$, where $\odot$ denotes pixel-wise product.

For the generator $G$, the final objective function is the combination of the adversarial loss $\mathcal{L}_{adv}$ and the reconstruction loss $\mathcal{L}_{\ell_1}$, which measures the coherence and similarity between predicted image and ground-truth image. Corresponding formulation can be expressed as
\begin{align}
    \label{Eq.gen_loss}
    \mathcal{L}_{G} & = \lambda_{adv} \cdot \mathcal{L}_{adv} + \lambda_{\ell_1} \cdot \mathcal{L}_{\ell_1} \\ \nonumber
    & = \lambda_{adv} \cdot \mathbb{E} [\log(1 - D(G(\boldsymbol{\mathrm{I}}_{in}, \boldsymbol{\mathrm{M}})))] + \lambda_{\ell_1} \cdot ||\boldsymbol{\mathrm{I}}_{out} - \boldsymbol{\mathrm{I}}_{gt}||_1,
\end{align}
where $\boldsymbol{\mathrm{I}}_{out}$ is the prediction of generator $G$, $\lambda_{adv}$ and $\lambda_{\ell_1}$ are the tradeoff parameters setting empirically.

%Considering the coherence and similarity with ground-truth images, the reconstruction loss of $G$ is defined as the $\ell_1$ distance between the output image of $G$ $\boldsymbol{\mathrm{I}}_{out} = G(\boldsymbol{\mathrm{I}}_{in}, \boldsymbol{\mathrm{M}})$ and the ground-truth image $\boldsymbol{\mathrm{I}}_{gt}$
%\begin{equation}
%    \label{Eq.l1_loss}
%    \mathcal{L}_{\ell_1} = ||\boldsymbol{\mathrm{I}}_{out} - \boldsymbol{\mathrm{I}}_{gt}||_1.
%\end{equation}
%For the generator $G$, the final minimize objective combined the adversarial loss and the reconstruction loss can be expressed as
%\begin{equation}
%    \label{Eq.gen_loss}
%    \mathcal{L}_{G} = \lambda_{adv} \cdot \mathcal{L}_{adv} + \lambda_{\ell_1} \cdot \mathcal{L}_{\ell_1},
%\end{equation}
%where $\lambda_{adv}$ and $\lambda_{\ell_1}$ are the tradeoff parameters setting empirically.

\subsection{Detection-based Generative Framework}
\label{subsec:detection}

As discussed above, the discriminator $D$ in GAN actually is a classifier, which just outputs a single scalar (label or probability). This scalar may not properly evaluate the quality of the generated image for image inpainting, where non-masked regions with much information easily has better reconstructed quality than the missing regions. The common average criterion in $D$ to some extent weakens this difference, more precisely, loses the exact position information of ``authentic'' artifacts. In extreme cases, such scalar might misguide the generator $G$ that pays less attention to the missing region.

%More importantly, it causes that $D$ output scalar is more correlative to value of mask ratio and less correlative to reconstructed quality, \textit{i.e.}, the larger mask ratio is, the higher value of output scalar will be.
%The generator $G$ reconstructed background is easier than the missing region.
%Different mask ratios for input corrupted images in dataset also become unstable during training period.

To solve the above problems, we proposed a novel detection-based generative framework for image inpainting. This framework consists of a generator $G$ and a detector $D_{et}$. The generator $G$ repairs the missing region to be harmony with contextual background. The detector $D_{et}$ evaluates the output $\boldsymbol{\mathrm{I}}_{out}$ of $G$ in pixel-wise manner and also localizes the unreasonable completion region, \textit{e.g.}, various artifacts, blurry patch, etc. Compared with a single scalar of $D$ in GAN, the precise localization significantly assists the generator $G$ in reconstructing of corrupted images.

%For the training of $D_{et}$, it is difficult to provide the ground-truth mask.
Unfortunately, it is difficult to provide the ground-truth mask for $D_{et}$. In this work, we adopt a weakly supervised strategy: we consider the binary mask $\boldsymbol{\mathrm{M}}$ as the ``proxy'' of the ground-truth mask. This is reasonable because the missing regions are usually difficult to predict than the non-masked regions (just reproducing the non-masked regions of the input image) and also have more artifacts. In addition, the similar pattern of corrupted regions in the non-masked regions are also captured by $D_{et}$ as the training progresses. The processing of this weakly supervised learning can be written as $\boldsymbol{\mathrm{V}} = D_{et}(\boldsymbol{\mathrm{I}}_{out})$, where $\boldsymbol{\mathrm{V}}$ is the output of detector $D_{et}$ to evaluate the prediction $\boldsymbol{\mathrm{I}}_{out}$. The valuation $\boldsymbol{\mathrm{V}}$ is the same size as the mask $\boldsymbol{\mathrm{M}}$, and its value, from 0 to 1, reflects the realistic degree of each pixel, \textit{i.e.}, the lower value indicates the more realistic completion result. Unlike the standard discriminator, the detector $D_{et}$ gives a human-like evaluation of the inpainted image, rather than a rough score. More analysis about the output of the detector $D_{et}$ will be shown in Section~\ref{subsec:visualization}.

%According to Eq.~\ref{Eq.semi_learning}, unlike the standard discriminator, the detector $D_{et}$ do not need to input the ground-truth image $\boldsymbol{\mathrm{I}}_{gt}$, which conveniently estimates the vraisemblance of $\boldsymbol{\mathrm{I}}_{out}$ during the test period, especially in the situation without $\boldsymbol{\mathrm{I}}_{gt}$.

%\begin{equation}
%    \label{Eq.semi_learning}
%    \boldsymbol{\mathrm{V}} = %D_{et}(\boldsymbol{\mathrm{I}}_{out}),
%\end{equation}

%Consistently, let $\boldsymbol{\mathrm{I}}_{gt}$ be the ground-truth image and $\boldsymbol{\mathrm{M}}$ be the binary mask (1 for holes). The generator $G$ uses corrupted image $\boldsymbol{\mathrm{I}}_{gt} \odot (1 - \boldsymbol{\mathrm{M}}) + \boldsymbol{\mathrm{M}}$ as the input $\boldsymbol{\mathrm{I}}_{in}$, and image mask $\boldsymbol{\mathrm{M}}$ as a pre-condition. Here, $\odot$ means pixel-wise product. Thus the output prediction $\boldsymbol{\mathrm{I}}_{out}$ of the generator $G$ is represented as $G(\boldsymbol{\mathrm{I}}_{in}, \boldsymbol{\mathrm{M}})$.

The coupled training should be executed between the generator $G$ and the detector $D_{et}$. As the coupled intermediate, $\boldsymbol{\mathrm{I}}_{out}$ is the input of the detector $D_{et}$, while valuation $\boldsymbol{\mathrm{V}}$, inserted in the reconstruction loss (see Eq.~\eqref{Eq.gen_loss}), constitutes a weighted reconstruction loss, which optimizes the generator $G$. A segmentation loss considering the imbalance between masked and non-masked regions is used to optimize the detector $D_{et}$. The above description is a typically adversarial process where the detector accurately locates artifacts, \textit{i.e.}, minimization of the segmentation-based loss, while the generator tries to deceive the detector that the probability of the artifacts in any location is the same, \textit{i.e.}, maximization of the segmentation-based loss. We will validate the proposed framework in Section~\ref{sec:exp_framework}, and discuss the details of the loss functions and their implementation in Section~\ref{sec:loss_function}.

%To avoid the distraction of the mask ratio and stabilize the training period, the focal loss is used to optimize the detector $D_{et}$.

\subsection{Network Architecture}
\label{sec:network}

The proposed image inpainting framework consists of two networks: the generator $G$ and the detector $D_{et}$.

\subsubsection{Generator}
The generator following an architecture similar to  EdgeConnect~\cite{nazeri2019edgeconnect} comprises three components: two down-sampling layers, eight residual blocks~\cite{he2016deep}, and two up-sampling layers, as shown in Fig.~\ref{Fig:Framework}. A convolutional layer with kernel size = 4 and stride = 2 executes down-sample, and up-sample is conducted by a deconvolutional layer with kernel size = 4 and stride = 2. Each of residual block holds two dilated convolutions~\cite{yu2015multi} with kernel size = 3 , stride = 1 and dilation factor = 2. After all convolutions/deconvolutions except the last convolution in whole network, instance normalization~\cite{ulyanov2017improved} and ReLU is followed, successively.
%the encoder that down-samples twice, eight residual blocks~\cite{he2016deep} that extract semantic features for recovery and the decoder that up-samples twice to return the original image size

\subsubsection{Detector}
The detector is a seven-layer fully convolutional network, which is augmented by in-network up-sampling and pixel-wise loss for dense evaluation of the inpainted image. The first five convolutional layers down-sample images twice, followed by two deconvolutional layers to up-sample images back to the original size. The final softmax layer transforms the output to the probability map $\boldsymbol{\mathrm{V}}$. Leaky ReLU with $\alpha = 0.2$ is used in down-sampling stage, and all convolutional kernel size is $4$.

%Due to the missing region of label 1 in the mask, valuation $\boldsymbol{\mathrm{V}}$ is the second channel of probability map.
%kernel size 4

\subsection{Loss Functions}
\label{sec:loss_function}

During the adversarial training, the generator targets both pixel-wise reconstruction precision and plausible visual result, while the detector aims at fine-grained evaluation for the prediction. We incorporate segmentation-based loss and weighted reconstruction loss to train our detector and generator, separately.

\subsubsection{Segmentation-based Loss}

%The cross entropy loss is the most common loss for image segmentation task. Our evaluation task implements pixel-wise binary cross entropy to localize unrealistic region, which can be described as
As described in Section~\ref{subsec:detection}, we use the weakly supervised learning to train our detector with the binary mask as ground-truth, therefore, a natural option is the standard pixel-wise binary cross entropy loss,

\begin{equation}
    \label{Eq.ce_loss}
    \mathcal{L}_{CE} = - \dfrac{1}{N} \sum_{i=1}^N \boldsymbol{\mathrm{M}}_i \log \boldsymbol{\mathrm{V}}_i + (1 - \boldsymbol{\mathrm{M}}_i) \log (1 - \boldsymbol{\mathrm{V}}_i),
\end{equation}
where $\boldsymbol{\mathrm{M}}_i$ and $\boldsymbol{\mathrm{V}}_i$ respectively are the one element of mask $\boldsymbol{\mathrm{M}}$ and valuation $\boldsymbol{\mathrm{V}}$, and $N$ is the number of elements in $\boldsymbol{\mathrm{M}}$. In Eq.~\eqref{Eq.ce_loss}, a smaller value of the loss function approximates more accurate valuation from the detector due to the weak supervision. Generally, the missing region in input image is smaller than the valid region, and this causes the imbalance between positive and negative samples. To balance the two classes, we introduce a weight $\alpha$, which is the mask ratio for input image. The balanced version of $\mathcal{L}_{CE}$ is formulated as
\begin{equation}
    \label{Eq.bce_loss}
    \mathcal{L}_{BCE} = - \dfrac{1}{N} \sum_{i=1}^N (1 - \alpha)\boldsymbol{\mathrm{M}}_i \log \boldsymbol{\mathrm{V}}_i + \alpha (1 - \boldsymbol{\mathrm{M}}_i) \log (1 - \boldsymbol{\mathrm{V}}_i).
\end{equation}
%and stable the training of $D_{et}$

In addition, we also consider a recent segmentation loss, \textit{i.e.}, focal loss~\cite{lin2017focal}, which is an enhanced version of Eq.~\eqref{Eq.bce_loss} with tunable focusing parameter $\gamma \geq 0$,
\begin{equation}
    \begin{split}
    \label{Eq.focal_loss}
    \mathcal{L}_{Focal} &= - \dfrac{1}{N} \sum_{i=1}^N (1 - \alpha) (1 - \boldsymbol{\mathrm{V}}_i)^{\gamma} \boldsymbol{\mathrm{M}}_i \log \boldsymbol{\mathrm{V}}_i \\
    &+ \alpha \boldsymbol{\mathrm{V}}_i^{\gamma} (1 - \boldsymbol{\mathrm{M}}_i) \log (1 - \boldsymbol{\mathrm{V}}_i).
    \end{split}
\end{equation}

In our experiments, we compare these three optional loss functions for training the detector, and find that the focal loss can yield the best performance, as shown in Section~\ref{sec:ablation}.
%\weize{Check this equ, I exchange the order of $\alpha$ and $1-\alpha$}

%
%On the other hand, the balanced cross entropy loss normalizes the size of the mask, which mitigates the effect of different mask ratios. Our experimental results in Section~ show that the balanced cross entropy loss makes the coupled training stable, and  accelerates convergence.

\subsubsection{Weighted Reconstruction Loss}
The generator learns with two objectives: the realistic visual quality and the consistency with the ground-truth image. The traditional adversarial framework combines these objectives with two hyper-parameters shown in Eq.~\eqref{Eq.gen_loss}, but it is still multi-objective optimization essentially. Compared with the single-objective optimization, multi-objective optimization is often difficult, \textit{e.g.}, the balance between the maximum margin and the minimum error is a tough problem in Soft-Margin SVM algorithm~\cite{chen2004support}. More importantly, tradeoff parameters without the physical significance to explain reduces generalization in different inpainting cases, \textit{i.e.}, the dataset of face, scene and street view needs corresponding parameters. To address this problem, we refer to Boosting algorithm~\cite{drucker1993boosting}, the idea is to increase the weight of weak samples and decrease the weight of strong samples. For a single image inpainting, the valuation $\boldsymbol{\mathrm{V}}$ distinguishes the weak or strong pixels appropriately. Therefore, our proposed weighted reconstruction loss merging two objectives can be written as

\begin{equation}
    \label{Eq.weight_loss}
    \mathcal{L}_{w} = \dfrac{1}{N} \sum_{i=1}^N \boldsymbol{\mathrm{W}}_i \cdot ||\boldsymbol{\mathrm{I}}_{out}^i - \boldsymbol{\mathrm{I}}_{gt}^i||_1,
\end{equation}
where $\boldsymbol{\mathrm{W}}_i$, $\boldsymbol{\mathrm{I}}_{out}^i$, and $\boldsymbol{\mathrm{I}}_{gt}^i$ are the pixel-wise weight $\boldsymbol{\mathrm{W}}$, prediction $\boldsymbol{\mathrm{I}}_{out}$ and ground-truth $\boldsymbol{\mathrm{I}}_{gt}$, respectively. Note that the weight $\boldsymbol{\mathrm{W}}$ maps to the valuation $\boldsymbol{\mathrm{V}}$, called weight mapping (please see following Section~\ref{subsec:weight_mapping} for details). Minimization of the weighted reconstruction loss for the generator is to multiply the artifact by a smaller weight, which is just the opposite target of the detector, described in Eq.~\eqref{Eq.ce_loss}. Competition between the generator and detector drives to improve their capabilities until the artifacts are hard to be perceived by naked eyes or the detector, which is the fundamental purpose of image inpainting task.

\begin{figure*}[htp]
  \centering
  \includegraphics[width=0.95\textwidth]{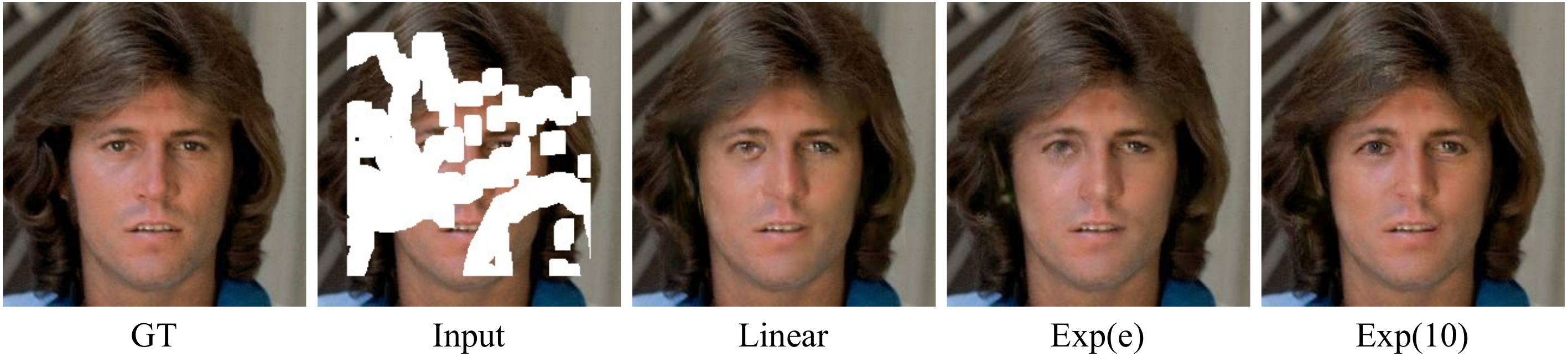}
  \caption{From left to right: the ground-truth image, the corrupted image, result using linear function, result using exponential function with e and 10 as base separately. The edge ratios after canny algorithm~\cite{canny1986computational} with $\sigma = 1.0$ are $10.14\%$ (GT), $8.61\%$ (Line), $8.72\%$ (Exp(e)) and $9.39\%$ (Exp(10)).}
  \label{Fig:weight}
\end{figure*}

\subsubsection{Weight Mapping}
\label{subsec:weight_mapping}
For high accuracy of the missing region reconstruction, we adopt to enhance the weight of weak pixels instead of subtracting the weight of strong pixels, so the range of the weight $\boldsymbol{\mathrm{W}}$ is $[1, +\infty)$. The weight mapping is a transition function from the valuation $[0, 1]$ to the weight $[1, +\infty)$, and the candidate is linear or exponential functions.
\begin{itemize}
    \item Linear transition can be written as  $\boldsymbol{\mathrm{W}} = 1 + \boldsymbol{\mathrm{V}}$, and the range of the weight $\boldsymbol{\mathrm{W}}$ is $[1, 2]$. Although the simple form is easy to implement, the low upper bound value causes less enhancement for awful reconstruction.
    \item Exponential transition can be written as  $\boldsymbol{\mathrm{W}} = x^{\boldsymbol{\mathrm{V}}}$, where $x$ ($x > 1$) is a base number of exponential function, and the range of the weight $\boldsymbol{\mathrm{W}}$ is $[1, x]$. Compared with linear transition, the exponential transition reduces the relative weight on well-inpainted pixels, putting more focus on artifacts.  Unlike the tradeoff parameters $\lambda_{adv}$ and $\lambda_{\ell_1}$, $x$ controls the refined texture completion in the missing region purposefully, as shown in Fig.~\ref{Fig:weight}. The larger value of the base number indicates the clearer and more exquisite edge or texture information.
\end{itemize}

To sum up, we combine the weighted reconstruction loss and focal loss as the final loss function to train the whole framework, which introduces the min-max adversarial process as follows:

\begin{equation}
    \label{Eq.final}
    \left\{
        \begin{array}{lr}
            \min\limits_G \; ||x^{Det(G(\boldsymbol{\mathrm{I}}_{in}, \boldsymbol{\mathrm{M}}))} \odot (G(\boldsymbol{\mathrm{I}}_{in}, \boldsymbol{\mathrm{M}}) - \boldsymbol{\mathrm{I}}_{gt})||_1, &  \\
            \max\limits_{Det} \; -\mathcal{L}_{Focal}(Det(G(\boldsymbol{\mathrm{I}}_{in}, \boldsymbol{\mathrm{M}})), \boldsymbol{\mathrm{M}}). &  
        \end{array}
    \right.
\end{equation}
From Eq.~\eqref{Eq.final}, the generator and detector in our framework just dispute about the weight value in the corrupted regions (not involved in non-masked regions), which is an improvement of the global competition in the GAN-based framework to solve image inpainting problem.
% where $\mathcal{L}_{CE}(\cdot)$ and $\mathcal{L}_{BCE}(\cdot)$ are alternatives to substitute $\mathcal{L}_{Focal}(\cdot)$.

%result using exponential function with $e$ as base number, and result using exponential function with 10 as base number.

%, as our experiments show in Section~.

\section{Experimental Results}

To validate our proposed method, we quantitatively and qualitatively compare our method with several recent state-of-the-art methods on three public datasets including CelebA-HQ~\cite{liu2015deep,karras2017progressive}, Places2~\cite{zhou2017places}, and Paris StreetView~\cite{doersch2012what}. Moreover, comparisons with relative inpainting frameworks under the same generator verify the effectiveness of our detection-based framework. Ablation study is conducted to choose the appropriate loss for the training of our detector, and the visualization of the valuation is also analyzed to validate the ability of our detector.

\subsection{Implementation Details}
\label{sec:implenmentation}

We first describe the details of three public datasets used in our experiments. CelebA-HQ~\cite{liu2015deep,karras2017progressive} is a high quality face dataset with 30K images, and we randomly select 27K images for training and the remaining 3K images for testing. For Places2~\cite{zhou2017places}, we select 30 categories for our training and testing. We randomly sample 2K images from per category in training split of Places2 to construct our training set (60K images). The corresponding 3K images in testing split of Places2 are directly as our testing set. For Paris StreetView~\cite{doersch2012what}, we keep original splits, \textit{i.e.}, 14,900 images for training and 100 images for testing.
%More than 10 million images in 365 indoor/outdoor scene categories constitute a challenging dataset Places2.

%Mask set is an important part for image inpainting experiments.
For the irregular training masks, we create 180K masks with/without border constraints from the source of QD-IMD~\cite{quick2018} that is a collection of 50 million human drawings. Several data augmentation operations: rotation, dilation, and cropping are adopted in sequence during the mask generation. The irregular mask set from Liu \textit{et al.}~\cite{liu2018image} including 12K masks is used as testing masks. Note that both training and testing masks are classified by ranges of the mask ratio from $[0.01, 0.1]$ to $[0.5, 0.6]$ with the step of $0.1$.

In view of computational cost, we resize images and masks to $256 \times 256$ as the input of networks. The Adam optimizer~\cite{kingma2014adam} is used to optimize the parameters of inpainting models with the learning rate of $10^{-4}$ and $\beta_1 = 0, \beta_2 = 0.9$. We train the model for 100 epochs with the batch size of $8$. During the training period, the focusing parameter $\gamma$ in focal loss (Eq.~\eqref{Eq.focal_loss}) is set to $2.0$, and the weight mapping mentioned in Section~\ref{subsec:weight_mapping} is set to the exponential transition with $10.0$ as the base number. 

\begin{figure*}[!htp]
  \centering
  \includegraphics[width=0.78\textwidth]{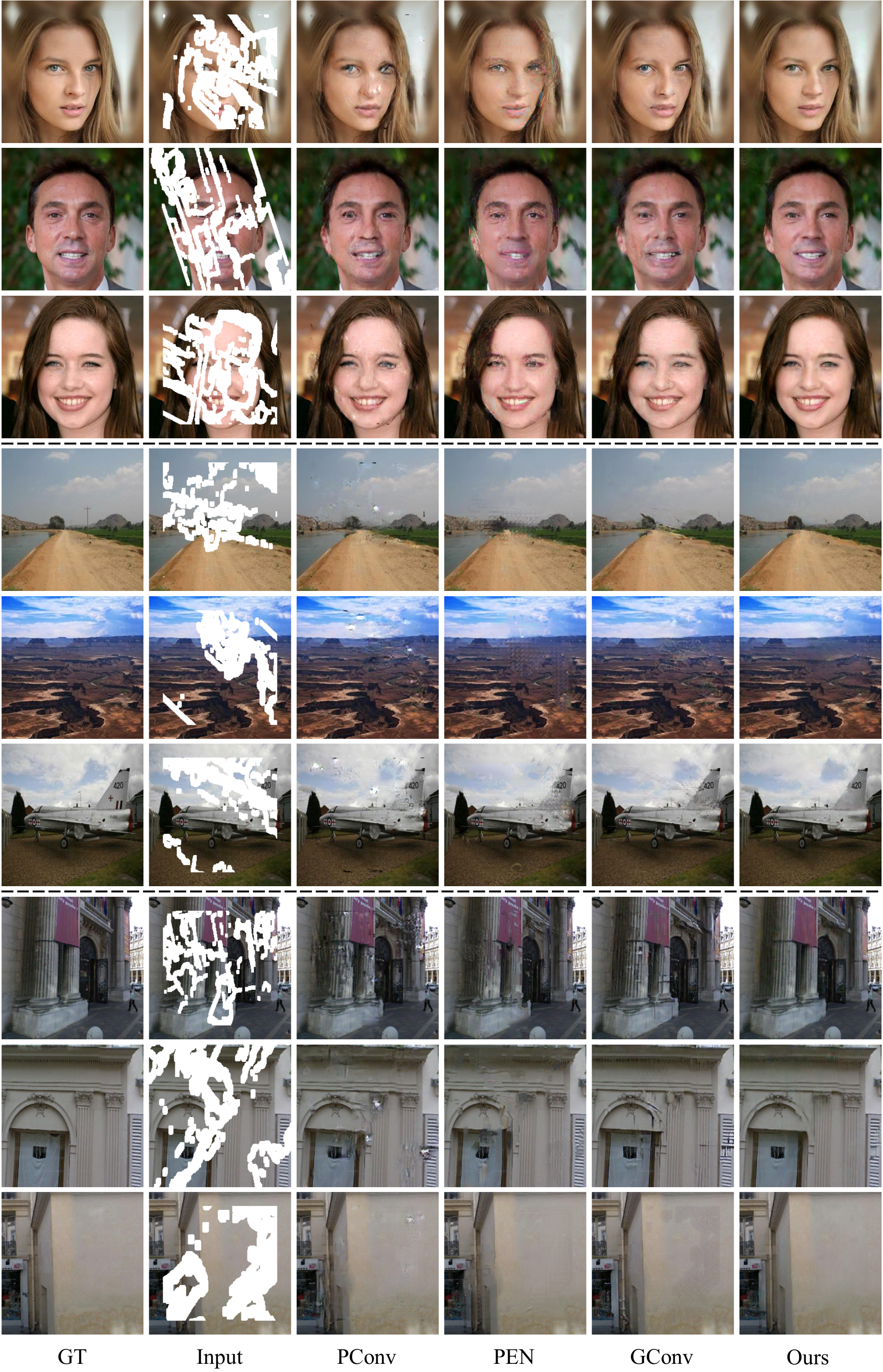}
  \caption{Example results of qualitative comparison. From top to bottom splited three groups from CelebA-HQ, Places2 and Paris StreetView testing set, respectively.}
  \label{Fig:comparison}
\end{figure*}

\begin{figure*}[htp]
  \centering
  \includegraphics[width=0.78\textwidth]{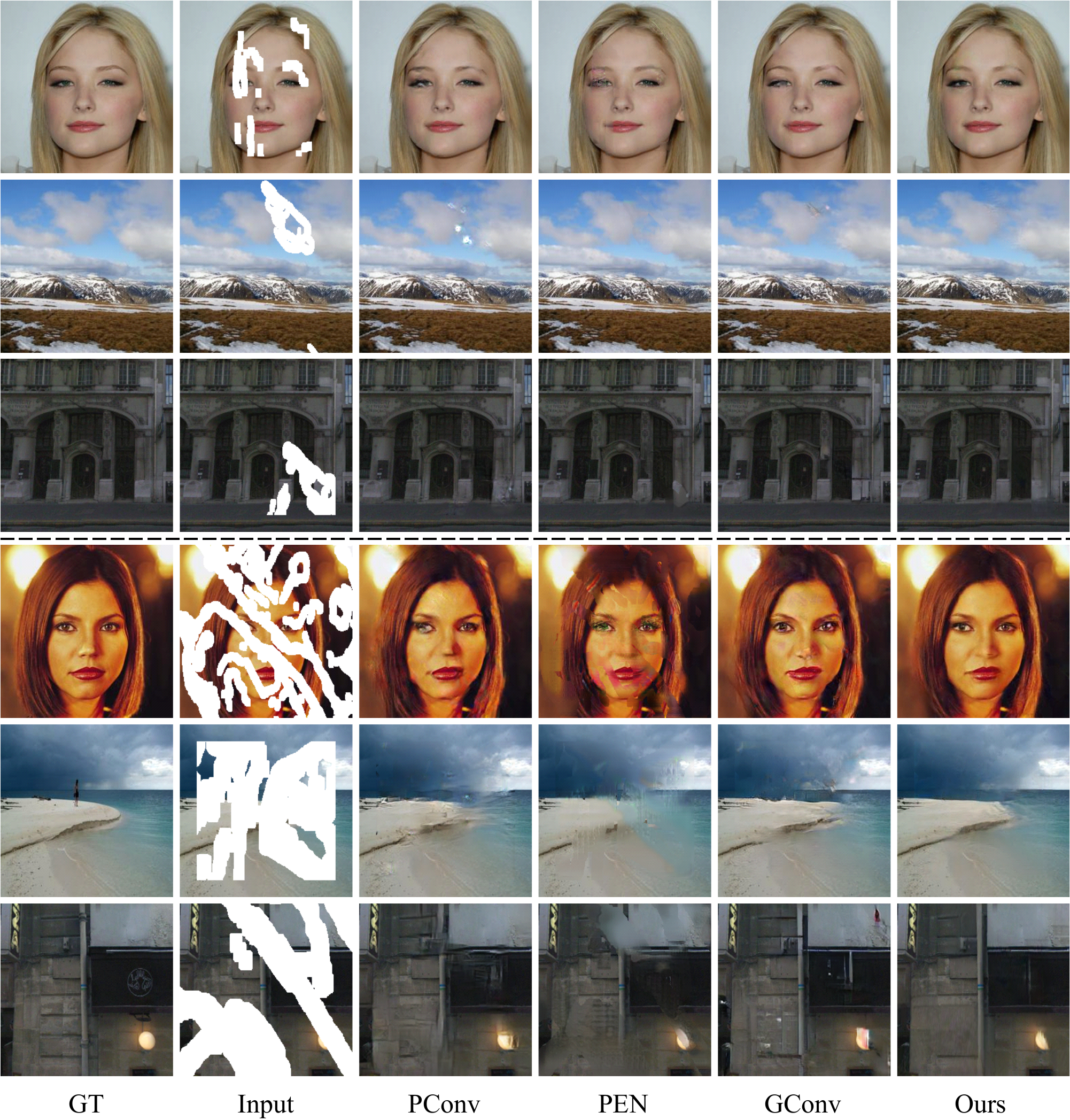}
  \caption{Example results for [0.01,  0.1] (top) and [0.5,  0.6] (bottom) mask from CelebA-HQ, Places2 and Paris StreetView testing set.}
  \label{Fig:extreme}
  %\vskip -0.5cm
\end{figure*}

\subsection{Comparison with State-of-the-arts}

We compare the proposed method with three representative state-of-the-art works that are different kinds of deep inpainting techniques:
\begin{itemize}
    \item PConv~\cite{liu2018image}: The method uses a novel partial convolution instead of standard convolution to solve inpainting problem.
    \item PEN~\cite{zeng2019learning}: A pyramid-context encoder network with a series of attention modules for inpainting.
    \item GConv~\cite{yu2019free}: A two-stage inpainting network with gated convolution, which is an upgrade of Contextual Attention~\cite{yu2018generative}.
\end{itemize}

For a fair comparison, these three existing methods and our method are all trained on three public datasets mentioned above. Officially released source codes of PEN and GConv are obtained from their respective project page~\cite{officail2019,deepfill2019}. As the source code of PConv is not available at the time of experiments, we use an unofficial implementation~\cite{unofficial2019} to train the model under our careful matching with their paper~\cite{liu2018image}. For each inpainting technique, the mean inference time of a $256 \times 256$ image on three datasets are recorded: PConv (27.97ms), PEN (61.08ms), GConv (18.91ms) and our method (18.80ms). Because of the one-stage network without special modules like attention layer leveraged in the generator, our method has an advantage on the time cost. Moreover, due to the short inference time of our detector (13.07ms), it is high-efficiency to train our whole framework.

\subsubsection{Qualitative Comparisons}

Fig.~\ref{Fig:comparison} shows some example outputs of four different models, \textit{i.e.}, PConv~\cite{liu2018image}, PEN~\cite{zeng2019learning}, GConv~\cite{yu2019free} and our method. There is no post-processing operation to ensure fairness. We observe that PConv~\cite{liu2018image} sometimes suffers from obvious visual artifacts and produces some meaningless textures. PEN~\cite{zeng2019learning} generates checkerboard artifacts in corrupted regions. It also shows poor coherence with background because of over-smoothing results and color inconsistency. GConv~\cite{yu2019free} produces better results but still exhibits imperfect details. Our method achieves more plausible results, especially in face image cases. More comparisons are provided in supplemental material.
In addition, we specifically illustrate some inpainting results on extreme mask range to verify our framework. As shown in Fig.~\ref{Fig:extreme}, for the trivial details, such as eyes and brows in face images, our results are slightly superior to that of other three methods on small mask range. Our method also achieves competitive results on large mask range in Fig.~\ref{Fig:extreme} and the following quantitative analysis.
%For all automatic inpainting 

%\begin{figure*}[htp]
%  \centering
%  \includegraphics[width=0.82\textwidth]{Large.pdf}
%  \caption{Example results for [0.5,  0.6] mask from CelebA-HQ, Places2 and Paris StreetView testing set.}
%  \label{Fig:large}
%\end{figure*}

%~\cite{karras2017progressive}    ~\cite{zhou2017places}

\begin{figure*}[t]
  \centering
  \includegraphics[width=0.8\textwidth]{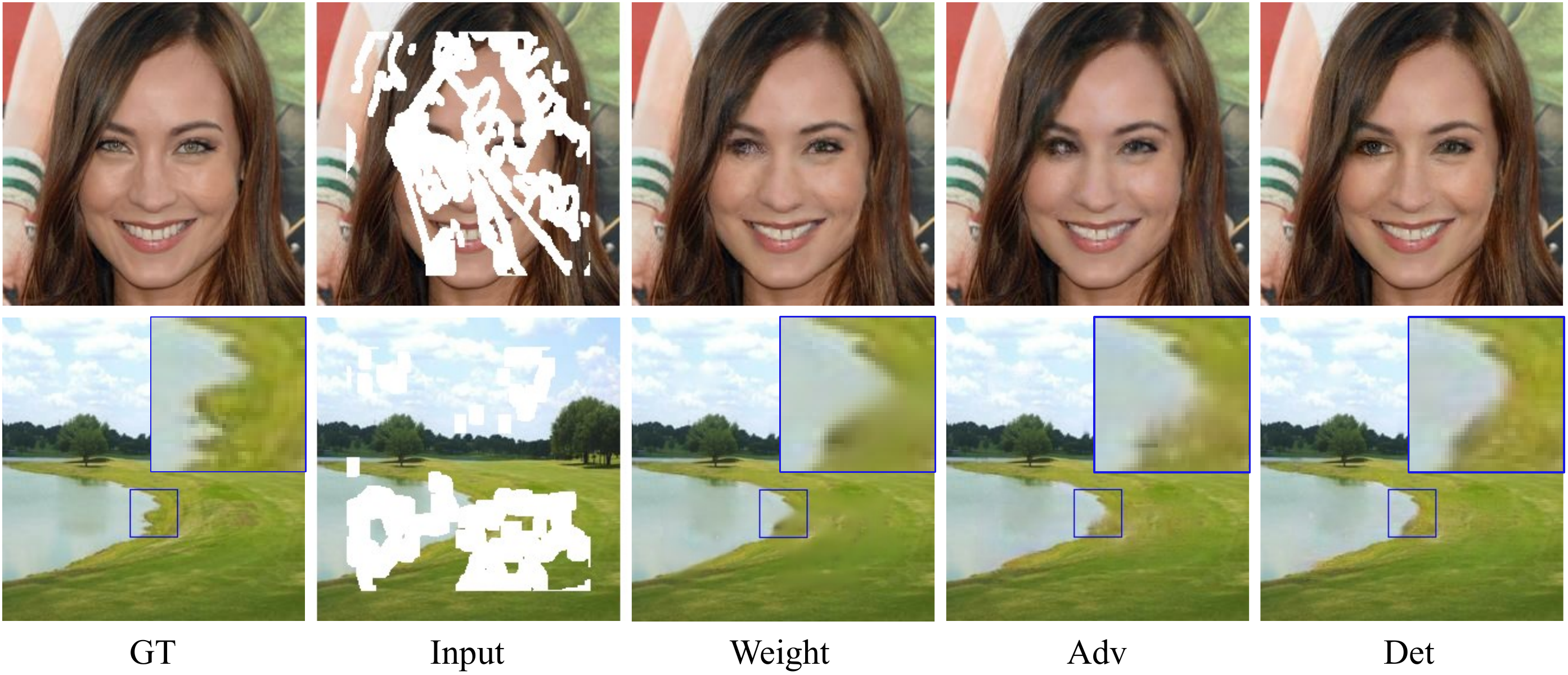}
  \caption{Comparison between results of hard-weighted $\ell_1$ loss based framework (Weight), typical adversarial framework (Adv) and detection-based generative framework (Det).}
  \label{Fig:comp_framework}
  \vskip -0.2cm
\end{figure*}

\setlength{\tabcolsep}{4pt}
\begin{table*}[t]
\begin{center}
\caption{Quantitative comparison of four different inpainting methods on CelebA-HQ and Places2. ``C'' stands for CelebA-HQ and ``P'' stands for Places2. Note that each statistic is calculated over all testing set in a fixed mask order. $\dagger$ Lower is better. $\P$ Higher is better.}
\label{Tab:comparison}
\footnotesize
\setlength{\tabcolsep}{5pt} \centering
%\resizebox{\textwidth}{!}{
  \begin{tabular}{c|c||c|c||c|c||c|c||c|c||c|c||c|c}
    \hline
    & Mask  & \multicolumn{2}{c||}{(0.01-0.1{]}} & \multicolumn{2}{c||}{(0.1-0.2{]}} & \multicolumn{2}{c||}{(0.2-0.3{]}} & \multicolumn{2}{c||}{(0.3-0.4{]}} & \multicolumn{2}{c||}{(0.4-0.5{]}} & \multicolumn{2}{c}{(0.5-0.6{]}}  \\ \hline
    & Data  & C              & P                 & C              & P                & C              & P                & C              & P                & C              & P                & C              & P               \\ \hline \hline
    \multirow{4}{*}{\rotatebox{90}{$\ell_1(\%)\dagger$}}
    & PConv & 0.84           & 1.27              & 2.05           & 3.07             & 3.57           & 5.16             & 5.43           & 7.37             & 7.59           & 9.94             & 12.00          & 14.14           \\ \cline{2-14}
    & PEN   & 0.80           & 1.11              & 2.15           & 2.96             & 3.88           & 5.25             & 5.83           & 7.59             & 8.02           & 10.19            & 11.77          & 13.95           \\ \cline{2-14}
    & GConv & \textbf{0.65}  & \textbf{1.00}     & 1.81           & 2.71             & 3.41           & 4.98             & 5.33           & 7.44             & 7.53           & 10.22            & 12.05          & 14.97           \\ \cline{2-14}
    & Ours  & 0.69           & \textbf{1.00}     & \textbf{1.62}  & \textbf{2.36}    & \textbf{2.82}  & \textbf{4.09}    & \textbf{4.28}  & \textbf{5.98}    & \textbf{6.06}  & \textbf{8.18}    & \textbf{10.05} & \textbf{12.10}  \\ \hline \hline
    \multirow{4}{*}{\rotatebox{90}{PSNR\P}}
    & PConv & 34.88          & 31.99             & 30.00          & 27.27            & 27.20          & 24.79            & 24.95          & 23.07            & 23.12          & 21.58            & 20.33          & 19.56           \\ \cline{2-14}
    & PEN   & 35.34          & 33.24             & 29.76          & 27.72            & 26.79          & 24.85            & 24.70          & 23.04            & 23.06          & 21.58            & 20.85          & 19.84           \\ \cline{2-14}
    & GConv & \textbf{37.14} & \textbf{34.05}    & 31.02          & 28.26            & 27.57          & 25.00            & 25.03          & 22.84            & 23.10          & 21.16            & 20.22          & 18.88           \\ \cline{2-14}
    & Ours  & 36.37          & 33.51             & \textbf{32.02} & \textbf{29.43}   & \textbf{29.13} & \textbf{26.64}   & \textbf{26.81} & \textbf{24.68}   & \textbf{24.89} & \textbf{23.04}   & \textbf{21.79} & \textbf{20.74}  \\ \hline \hline
    \multirow{4}{*}{\rotatebox{90}{SSIM\P}}
    & PConv & 0.983          & 0.961             & 0.963          & 0.917            & 0.936          & 0.866            & 0.898          & 0.809            & 0.851          & 0.738            & 0.744          & 0.615           \\ \cline{2-14}
    & PEN   & 0.988          & 0.975             & 0.965          & 0.929            & 0.933          & 0.867            & 0.894          & 0.801            & 0.849          & 0.723            & 0.764          & 0.605           \\ \cline{2-14}
    & GConv & \textbf{0.991} & \textbf{0.978}    & 0.971          & 0.936            & 0.941          & 0.876            & 0.902          & 0.808            & 0.856          & 0.73             & 0.750          & 0.594           \\ \cline{2-14}
    & Ours  & 0.990          & 0.977             & \textbf{0.977} & \textbf{0.949}   & \textbf{0.958} & \textbf{0.907}   & \textbf{0.930} & \textbf{0.856}   & \textbf{0.894} & \textbf{0.792}   & \textbf{0.793} & \textbf{0.665}  \\ \hline \hline
    \multirow{4}{*}{\rotatebox{90}{FID$\dagger$}}
    & PConv & 2.79           & 4.44              & 4.42           & 8.38             & 5.60           & 12.7             & 7.73           & 17.46            & 11.02          & 24.3             & 15.16          & \textbf{32.75}  \\ \cline{2-14}
    & PEN   & 1.41           & 3.1               & 4.19           & 8.76             & 8.38           & 17.31            & 12.68          & 29.84            & 18.73          & 47.05            & 23.38          & 66.7            \\ \cline{2-14}
    & GConv & \textbf{0.78}  & \textbf{2.27}     & 2.05           & 6.02             & 3.93           & 11.02            & 5.86           & 16.39            & 8.64           & \textbf{22.75}   & \textbf{12.75} & 33.17           \\ \cline{2-14}
    & Ours  & 1.08           & 2.82              & \textbf{1.86}  & \textbf{5.78}    & \textbf{3.34}  & \textbf{10.38}   & \textbf{5.16}  & \textbf{16.18}   & \textbf{7.84}  & 23.89            & 15.34          & 37.18           \\ \hline
  \end{tabular}
  %}
\end{center}
\end{table*}

\subsubsection{Quantitative Comparisons}

Multiple reasonable contents combined with contextual background constitute a realistic image, which may be different from the ground-truth image. Because the nature of the non-unique solution of image inpainting problem, numerical metrics are difficult to evaluate the quality of a single inpainting case. However, the mean value of metrics on whole dataset could measure the performance of inpainting techniques. In our work, we follow the previous inpainting works and measure the results with four metrics: $\ell_1$ error, peak signal-to-noise ratio (PSNR), structural similarity index (SSIM)~\cite{wang2004image} and Fr{\'e}chet Inception Distance (FID)~\cite{heusel2017gans}. The first three metrics calculate pixel-wise deviation under the assumption that recovery regions target to the ground-truth images. FID based on semantic measurement scales the Wasserstein-2 distance between distributions of real and reconstructed images with a pre-trained Inception-V3 model~\cite{szegedy2016rethinking}. Table~\ref{Tab:comparison} reports the quantitative comparison results of all four methods on CelebA-HQ and Places2 dataset. Our method performs best for all measure in the range $[0.1, 0.4]$, and we could still achieve competitive results with other works in too low/too high intervals. As the detector is not sensitive to the mask with much small mask ratio, our model has less poor performance at the mask range of $[0.01, 0.1]$ , compared with GConv model.

\begin{figure*}[t]
  \centering
  \includegraphics[width=0.8\textwidth]{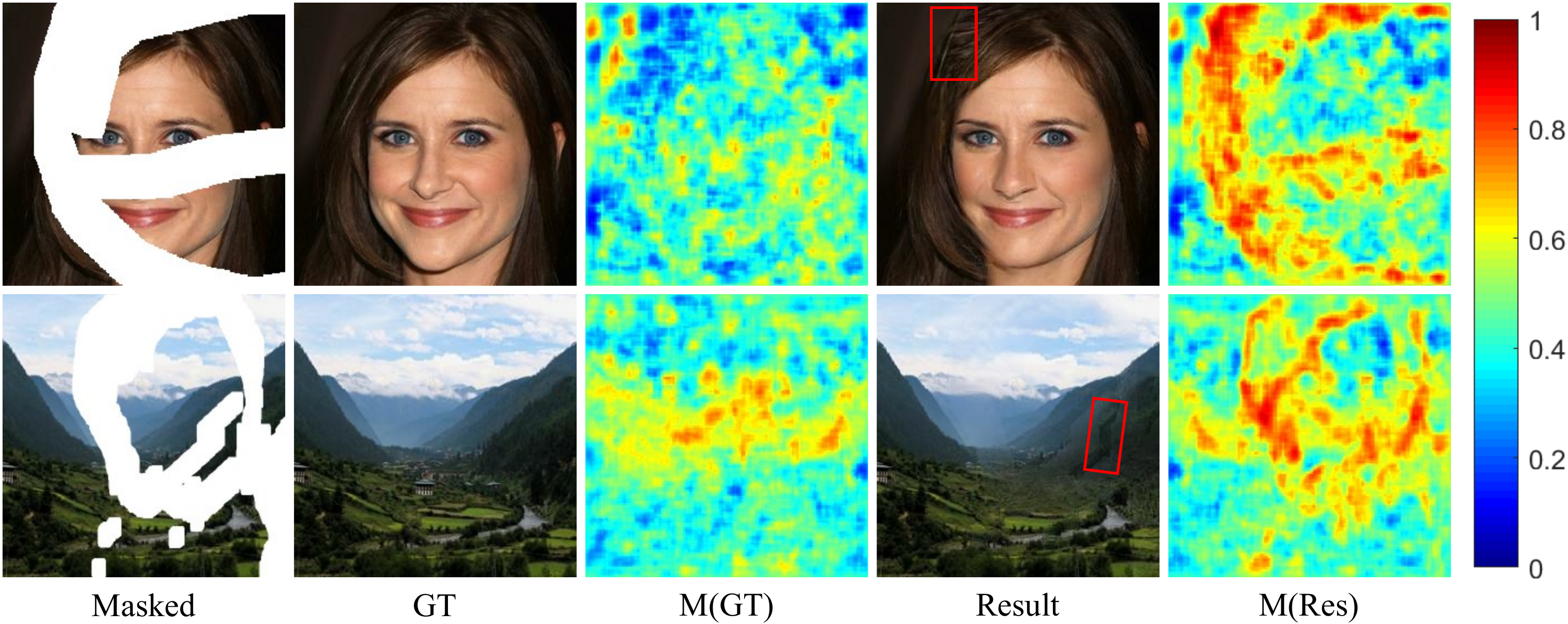}
  \caption{Visualization of the output of the detector. }
  \label{Fig:Visualization}
  \vskip -0.2cm
\end{figure*}

\setlength{\tabcolsep}{4pt}
\begin{table*}[t]
\begin{center}
\caption{The evaluation results of ablation study about loss function of detector: the cross entropy loss (CE), the balanced cross entropy loss (BCE), and the focal loss (Focal).  Higher is better for both two metrics.}
\label{Tab:ablation}
\footnotesize
\setlength{\tabcolsep}{5pt} \centering
%\resizebox{\textwidth}{!}{
  \begin{tabular}{c||c|c||c|c||c|c||c|c||c|c||c|c}
    \hline
    Mask     & \multicolumn{2}{c||}{(0.01-0.1{]}} & \multicolumn{2}{c||}{(0.1-0.2{]}} & \multicolumn{2}{c||}{(0.2-0.3{]}} & \multicolumn{2}{c||}{(0.3-0.4{]}} & \multicolumn{2}{c||}{(0.4-0.5{]}} & \multicolumn{2}{c }{(0.5-0.6{]}} \\ \hline
    Metrics  & PSNR            & SSIM            & PSNR            & SSIM           & PSNR            & SSIM           & PSNR            & SSIM           & PSNR            & SSIM           & PSNR            & SSIM           \\ \hline \hline
    CE       & 35.88           & 0.989           & 31.77           & 0.976          & 28.9            & 0.956          & 26.56           & 0.927          & 24.64           & 0.890          & 21.49           & 0.784          \\ \hline
    BCE & 36.29           & 0.990           & 31.88           & 0.976          & 28.95           & 0.956          & 26.6            & 0.927          & 24.67           & 0.889          & 21.49           & 0.784          \\ \hline
    Focal    & \textbf{36.37}  & \textbf{0.990}  & \textbf{32.02}  & \textbf{0.977} & \textbf{29.13}  & \textbf{0.958} & \textbf{26.81}  & \textbf{0.93}  & \textbf{24.89}  & \textbf{0.894} & \textbf{21.79}  & \textbf{0.793} \\ \hline
\end{tabular}
%}
\end{center}
\end{table*}

\subsection{Comparisons with Relative Inpainting Frameworks}
\label{sec:exp_framework}

We also compare the proposed detection-based generative framework with two relative image inpainting frameworks:
\begin{itemize}
  \item Weight: The framework just involves the generator without the discriminator. Its objective function usually adopts hard-weighted $\ell_1$ loss, \textit{i.e.}, assign heavy weight ($\lambda_{1}$) to the corrupted regions, and light weight ($\lambda_{2}$) to non-masked regions. We set $\lambda_1 = 6$ and $\lambda_2 = 1$, which is the same as PConv~\cite{liu2018image}.
  \item Adv: The framework, first used by Context Encoders~\cite{pathak2016context}, combines reconstruction loss with adversarial loss to train the model. Most recent deep inpainting methods including PEN~\cite{zeng2019learning} and GConv~\cite{yu2019free} adopts it as well.
  \item Det: The detection-based generative framework we proposed applies in image inpainting task.
\end{itemize}

For a fair comparison, the generator in above three frameworks are the same, and detailed structure is described in Section~\ref{sec:network}. Sampled results of different frameworks are shown in Fig.~\ref{Fig:comp_framework}. In the first row, Weight and Adv fail to reconstruct reasonable structures of eyes. As we see zoom-in regions, Weight produces the over-smoothing result at the border of bank and lake, and the result of Adv is also blurry. Det outperforms all the other frameworks in detail, it is largely because of the advanced competition about ``Where are the artifacts?'' instead of ``Is the image real or fake?''.

\subsection{Ablation Study}
\label{sec:ablation}

In this experiment, we conduct an ablation study to evaluate the performance of several optional detection loss on CelebA-HQ, \textit{i.e.}, the standard cross entropy loss, the balanced loss, and the focal loss. We use the six ranges of the mask ratio mentioned in Section~\ref{sec:implenmentation}. The corresponding results are reported in Table~\ref{Tab:ablation}. We find that the focal loss finally achieves the highest performance, especially for PSNR. Therefore, we choose the focal loss as our loss function to train the detector. However, the results of focal loss are slightly better than that of standard cross entropy loss. The reason is that although the focal loss can solve imbalance problem to improve the detection accuracy~\cite{lin2017focal}, but such improvement implicitly affect the final inpainting results.

%The result is reasonable because the focal loss eliminates the class imbalance via mask-ratio-based balance strategy and puts more focus on hard, misclassified examples.

%The reason to involve the this type of weighted loss is to avoid the imbalance of samples between the corrupted regions and background.
%Although the standard cross entropy loss can already achieve good performance (comparing the rows of ``CE'' in Table~\ref{Tab:ablation} with ``PConv'', ``PEN'', and ``GConv'' in the columns of ``C'' in Table~\ref{Tab:comparison}), which to some extent illustrates the effectiveness of our detection-based generative framework,

\subsection{Visualization of Detection}
\label{subsec:visualization}

The detector is an important component in our proposed framework, and its results are close related to the inpainting quality. Through visualizing the evaluation of the inpainted images and the corresponding ground-truth images, we analyze how the detector perceives the artifacts by weakly supervised learning. In detail, we respectively feed the inpainted image and corresponding ground-truth image into the trained detector, and then visualize the colormaps of the output of the detector. The visualization results are shown in Fig.~\ref{Fig:Visualization}, where red color means the region has artifacts with high probability, and blue color means low probability. We find that the colormaps of the ground-truth images (``M(GT)'' in Fig.~\ref{Fig:Visualization}) are irregular showing a bit relation with image contents, whereas the colormaps of inpainted images (``M(Res)'' in Fig.~\ref{Fig:Visualization}) show strong correlation with input masks (``Masked'' in Fig.~\ref{Fig:Visualization}), \textit{i.e.}, most of red pixels are in/beside the mask regions. In the meanwhile, the obvious visual artifacts (the regions highlighting with the red rectangle in inpainted results) match with the hottest regions in the colormaps of inpainted results. Specially, in the second row of Fig.~\ref{Fig:Visualization}, the red rectangle in ``Result'' corresponds to the non-masked region around the missing region in ``Masked''. This means the detector learns the location of the artifacts rather than the binary mask. Consequently, the output of detector has the ability to indicate the position of defects, which is to some extent consistent with the perception of human eyes, and this useful position information is inserted into reconstruction loss to enhance visual artifacts.

\begin{figure*}[htp]
  \centering
  \includegraphics[width=0.95\textwidth]{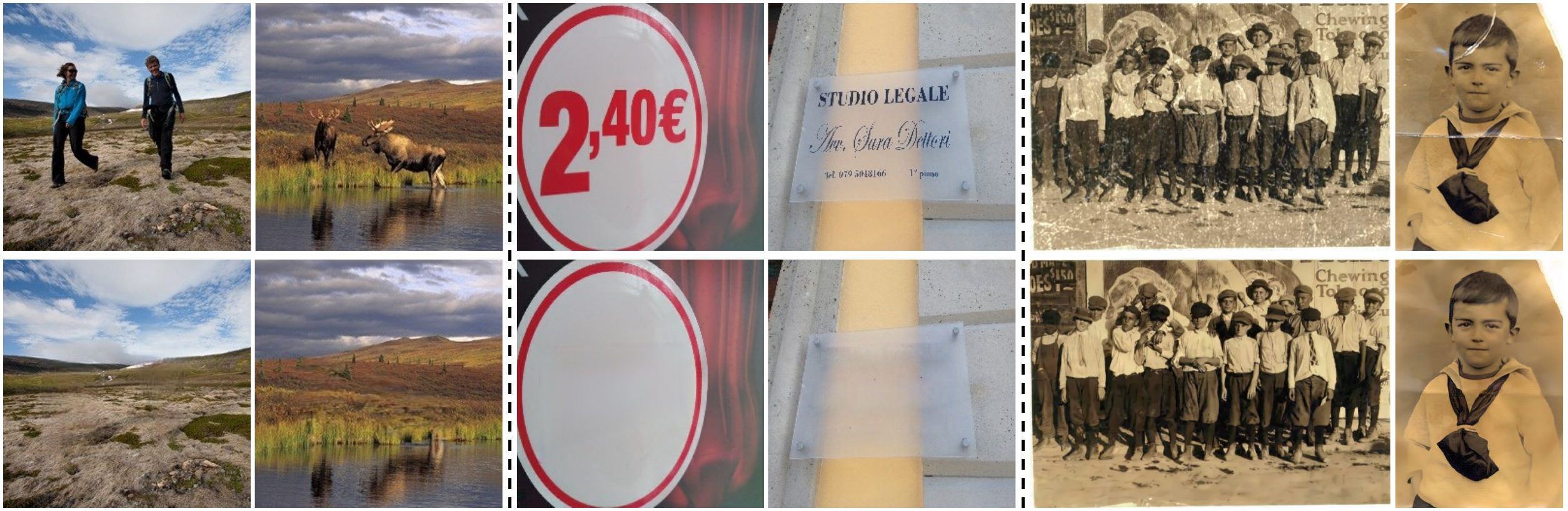}
  \caption{Example results of daily applications. From left to right, three split groups are results of object removal, text removal and old photo restoration, respectively. For each pair of images, the top image is the input and the bottom image is the image translation result.}
  \label{Fig:application}
  %\vskip -0.2cm
\end{figure*}

\subsection{Real-world Applications}

We demonstrate some daily applications of our whole framework on image translation. Fig.~\ref{Fig:application} shows our results of object removal (the left column), text removal (the middle column) and old photo restoration (the right column). We leverage the model trained on Places2 dataset without fine-tuning to conduct object removal task, which appears the harmonious filling contents, especially at the mask boundaries. Due to the large domain gaps between textual images and inpainting benchmarks, we retrain our model on the dataset collected in real world to solve text removal problem. The experimental results show that our method can handle the complex illumination and noise in real scenarios. Following \cite{wan2020bringing}, we synthesize amount of training data to train the model and then repair the old photos, and the results indicate that our method can recover unstructured degradation and structured scratches.

\section{Discussion and Conclusion}

\begin{figure}[ht]
  \centering
  \includegraphics[width=0.48\textwidth]{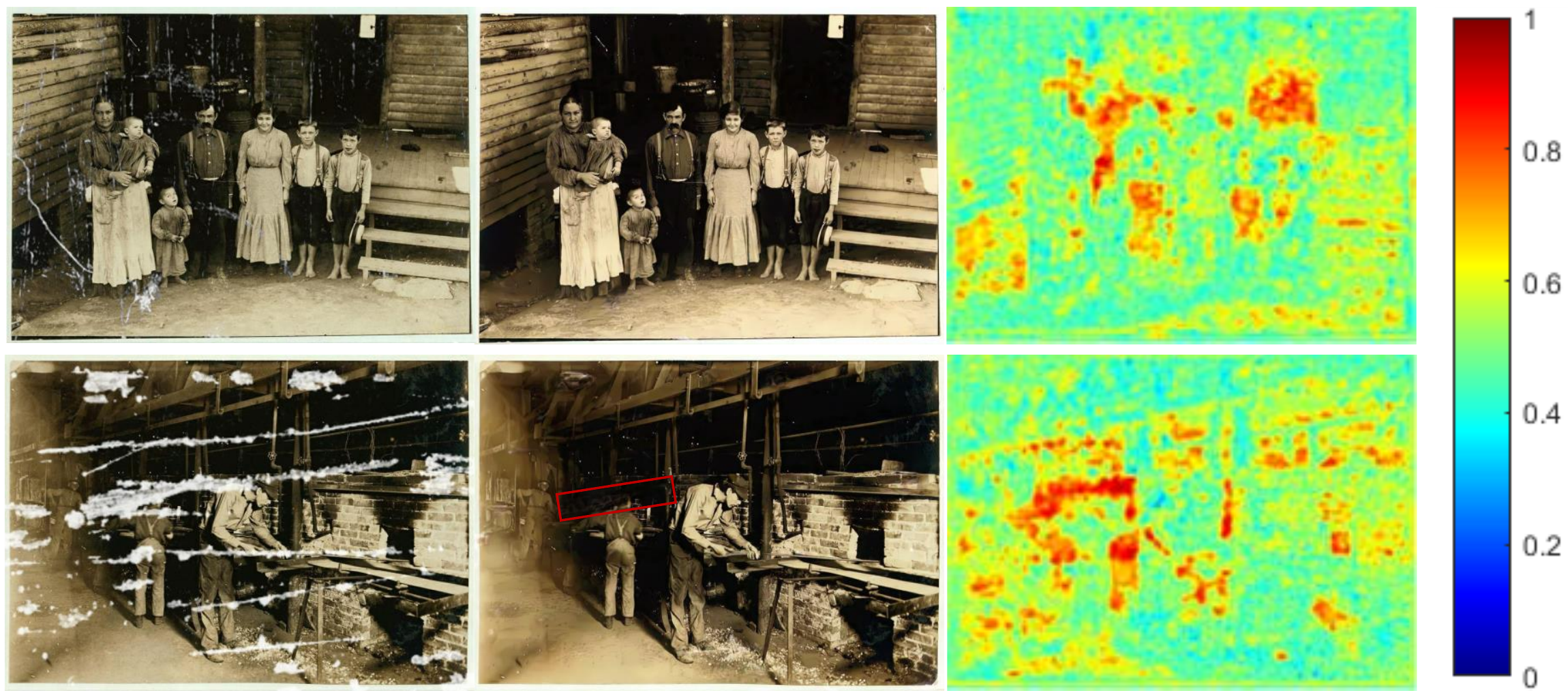}
  \caption{Visualization of the old photo restoration. From left to right are input images, repaired results and visualizations of repaired results, respectively.}
  \label{Fig:limitation}
  %\vskip -0.2cm
\end{figure}

In this paper, we proposed a detection-based generative framework to address image inpainting problem. To perceptually localize visual artifacts in inpainted images, we introduced a dense detector, which is trained by weakly supervised learning, to evaluate the quality of inpainted images in a pixel-wise manner. Furthermore, the reconstruction loss is combined with such evaluation using a weighting criterion to train the generator, which avoid tuning the tradeoff parameters manually. Extensive experiments demonstrate the superiority of proposed detection-based image inpainting framework. However, semantic information may disturb our detector to omit feeble artifacts existing in inpainted images during training period. From Table~\ref{Tab:comparison}, this obstruction is apparent in case of small mask ratio. As Fig.~\ref{Fig:limitation} shown, the detector easily captures artifacts from heavy scratches (the region highlighting with the red rectangle) in the second row, while the detector pays more attention to semantic information instead of tiny scratches in the first row. For large mask ratio, mask with too rough information of artifacts location may not be suitable to weakly supervised learning of the detector. Accurate artifacts localization is an open problem for the detection-based framework to solve. In future work, this framework may extend to other conditional generative tasks, \textit{e.g.}, image synthesis and image denoising. We also plan to implement our approach with the open source platform,  Jittor~\cite{hu2020Jittor}, for shorter training time.

%We demonstrated that such evaluation with position information of artifacts significantly improve inpainting results. 

\section*{Acknowledgements} This work was partially supported by the National Key R\&D Program of China (2019YFB2204104), the National Natural Science Foundation of China (61772523),  the Beijing Natural Science Foundation (L182059), the Tencent AI Lab Rhino-Bird Focused Research Program (No. JR202023), and the Open Research Fund Program of State key Laboratory of Hydroscience and Engineering, Tsinghua University (sklhse-2020-D-07).

%-------------------------------------------------------------------------
% bibtex
%\bibliographystyle{eg-alpha}
%\bibliography{egbibsample}        

% biblatex with biber
\printbibliography                

@string{NIPS = "Advances in Neural Information Processing Systems"}

@string{ICCV = "Proceedings of the IEEE international conference on computer vision"}

@string{CVPR = "Proceedings of the IEEE conference on computer vision and pattern recognition"}

@string{ECCV = "European Conference on Computer Vision"}

@string{SIGGRAPH = "Proceedings of the annual conference on Computer Graphics and Interactive Techniques"}

@string{TOG = "ACM Transactions on Graphics"}

@string{PAMI = "IEEE Transactions on Pattern Analysis and Machine Intelligence"}

@string{TIP = "IEEE Transactions on Image Processing"}

@string{ICCV = "{IEEE} {ICCV}"}

@string{CVPR = "{IEEE} {CVPR}"}

@string{ECCV = "{ECCV}"}

@string{SIGGRAPH = "ACM Trans. Graph. (Proc. {SIGGRAPH})"}

@string{SIGGRAPH_OLD = "Proc. ACM {SIGGRAPH}"}

@string{TOG = "ACM Trans. Graph."}

@string{PAMI = "{IEEE} Trans. Pattern Anal. Mach. Intell."}

@string{TIP = "{IEEE} Trans. Image Process."}

@inproceedings{goodfellow2014generative,
  title={Generative adversarial nets},
  author={Goodfellow, Ian and Pouget-Abadie, Jean and Mirza, Mehdi and Xu, Bing and Warde-Farley, David and Ozair, Sherjil and Courville, Aaron and Bengio, Yoshua},
  booktitle=NIPS,
  pages={2672--2680},
  year={2014}
}

@article{nazeri2019edgeconnect,
  title={{EdgeConnect}: Generative image inpainting with adversarial edge learning},
  author={Nazeri, Kamyar and Ng, Eric and Joseph, Tony and Qureshi, Faisal Z and Ebrahimi, Mehran},
  journal={arXiv preprint arXiv:1901.00212},
  year={2019}
}

@inproceedings{long2015fully,
  title={Fully convolutional networks for semantic segmentation},
  author={Long, Jonathan and Shelhamer, Evan and Darrell, Trevor},
  booktitle=CVPR,
  pages={3431--3440},
  year={2015}
}

@inproceedings{lin2017focal,
  title={Focal loss for dense object detection},
  author={Lin, Tsung-Yi and Goyal, Priya and Girshick, Ross and He, Kaiming and Doll{\'a}r, Piotr},
  booktitle=ICCV,
  pages={2980--2988},
  year={2017}
}

@inproceedings{he2016deep,
  title={Deep residual learning for image recognition},
  author={He, Kaiming and Zhang, Xiangyu and Ren, Shaoqing and Sun, Jian},
  booktitle=CVPR,
  pages={770--778},
  year={2016}
}

@inproceedings{ulyanov2017improved,
  title={Improved texture networks: Maximizing quality and diversity in feed-forward stylization and texture synthesis},
  author={Ulyanov, Dmitry and Vedaldi, Andrea and Lempitsky, Victor},
  booktitle=CVPR,
  pages={6924--6932},
  year={2017}
}

@article{chen2004support,
  title={Support vector machine soft margin classifiers: error analysis},
  author={Chen, Di-Rong and Wu, Qiang and Ying, Yiming and Zhou, Ding-Xuan},
  journal={Journal of Machine Learning Research},
  volume={5},
  number={Sep},
  pages={1143--1175},
  year={2004}
}

@incollection{drucker1993boosting,
  title={Boosting performance in neural networks},
  author={Drucker, Harris and Schapire, Robert and Simard, Patrice},
  booktitle={Advances in Pattern Recognition Systems using Neural Network Technologies},
  pages={61--75},
  year={1993},
  publisher={World Scientific}
}

@inproceedings{bertalmio2000image,
  title={Image inpainting},
  author={Bertalmio, Marcelo and Sapiro, Guillermo and Caselles, Vincent and Ballester, Coloma},
  booktitle=SIGGRAPH_OLD,
  pages={417--424},
  year={2000}
}

@inproceedings{efros2001image,
  title={Image quilting for texture synthesis and transfer},
  author={Efros, Alexei A and Freeman, William T},
  booktitle=SIGGRAPH_OLD,
  pages={341--346},
  year={2001}
}

@article{ballester2001filling,
  title={Filling-in by joint interpolation of vector fields and gray levels},
  author={Ballester, Coloma and Bertalmio, Marcelo and Caselles, Vicent and Sapiro, Guillermo and Verdera, Joan},
  journal=TIP,
  volume={10},
  number={8},
  pages={1200--1211},
  year={2001}
}

@article{barnes2009patchmatch,
  title={{PatchMatch}: A randomized correspondence algorithm for structural image editing},
  author={Barnes, Connelly and Shechtman, Eli and Finkelstein, Adam and Goldman, Dan B},
  journal=TOG,
  volume={28},
  number={3},
  pages={24},
  year={2009}
}

@inproceedings{simakov2008summarizing,
  title={Summarizing visual data using bidirectional similarity},
  author={Simakov, Denis and Caspi, Yaron and Shechtman, Eli and Irani, Michal},
  booktitle=CVPR,
  pages={1--8},
  year={2008}
}

@article{darabi2012image,
  title={{Image Melding}: combining inconsistent images using patch-based synthesis},
  author={Darabi, Soheil and Shechtman, Eli and Barnes, Connelly and Goldman, Dan B and Sen, Pradeep},
  journal=SIGGRAPH,
  volume={31},
  number={4},
  pages={1--10},
  year={2012}
}

@article{huang2014image,
  title={Image completion using planar structure guidance},
  author={Huang, Jia-Bin and Kang, Sing Bing and Ahuja, Narendra and Kopf, Johannes},
  journal=SIGGRAPH,
  volume={33},
  number={4},
  pages={1--10},
  year={2014}
}

@inproceedings{pathak2016context,
  title={Context encoders: Feature learning by inpainting},
  author={Pathak, Deepak and Krahenbuhl, Philipp and Donahue, Jeff and Darrell, Trevor and Efros, Alexei A},
  booktitle=CVPR,
  pages={2536--2544},
  year={2016}
}

@inproceedings{yang2017high,
  title={High-resolution image inpainting using multi-scale neural patch synthesis},
  author={Yang, Chao and Lu, Xin and Lin, Zhe and Shechtman, Eli and Wang, Oliver and Li, Hao},
  booktitle=CVPR,
  pages={6721--6729},
  year={2017}
}

@article{iizuka2017globally,
  title={Globally and locally consistent image completion},
  author={Iizuka, Satoshi and Simo-Serra, Edgar and Ishikawa, Hiroshi},
  journal=SIGGRAPH,
  volume={36},
  number={4},
  pages={1--14},
  year={2017}
}

@inproceedings{yu2018generative,
  title={Generative image inpainting with contextual attention},
  author={Yu, Jiahui and Lin, Zhe and Yang, Jimei and Shen, Xiaohui and Lu, Xin and Huang, Thomas S},
  booktitle=CVPR,
  pages={5505--5514},
  year={2018}
}

@inproceedings{zeng2019learning,
  title={Learning pyramid-context encoder network for high-quality image inpainting},
  author={Zeng, Yanhong and Fu, Jianlong and Chao, Hongyang and Guo, Baining},
  booktitle=CVPR,
  pages={1486--1494},
  year={2019}
}

@inproceedings{liu2019coherent,
  title={Coherent semantic attention for image inpainting},
  author={Liu, Hongyu and Jiang, Bin and Xiao, Yi and Yang, Chao},
  booktitle=ICCV,
  pages={4170--4179},
  year={2019}
}

@inproceedings{liu2018image,
  title={Image inpainting for irregular holes using partial convolutions},
  author={Liu, Guilin and Reda, Fitsum A and Shih, Kevin J and Wang, Ting-Chun and Tao, Andrew and Catanzaro, Bryan},
  booktitle=ECCV,
  pages={85--100},
  year={2018}
}

@inproceedings{yu2019free,
  title={Free-form image inpainting with gated convolution},
  author={Yu, Jiahui and Lin, Zhe and Yang, Jimei and Shen, Xiaohui and Lu, Xin and Huang, Thomas S},
  booktitle=ICCV,
  pages={4471--4480},
  year={2019}
}

@inproceedings{ren2019structureflow,
  title={StructureFlow: Image Inpainting via Structure-aware Appearance Flow},
  author={Ren, Yurui and Yu, Xiaoming and Zhang, Ruonan and Li, Thomas H and Liu, Shan and Li, Ge},
  booktitle=ICCV,
  pages={181--190},
  year={2019}
}

@article{simonyan2014very,
  title={Very deep convolutional networks for large-scale image recognition},
  author={Simonyan, Karen and Zisserman, Andrew},
  journal={arXiv preprint arXiv:1409.1556},
  year={2014}
}

@inproceedings{johnson2016perceptual,
  title={Perceptual losses for real-time style transfer and super-resolution},
  author={Johnson, Justin and Alahi, Alexandre and Fei-Fei, Li},
  booktitle=ECCV,
  year={2016},
  pages={694-711}
}

@InProceedings{Gatys_2016_image,
    author={Gatys, Leon A. and Ecker, Alexander S. and Bethge, Matthias},
    title={Image Style Transfer Using Convolutional Neural Networks},
    booktitle=CVPR,
    pages={2414-2423},
    year={2016}
}

@inproceedings{xie2019image,
  title={Image Inpainting with Learnable Bidirectional Attention Maps},
  author={Xie, Chaohao and Liu, Shaohui and Li, Chao and Cheng, Ming-Ming and Zuo, Wangmeng and Liu, Xiao and Wen, Shilei and Ding, Errui},
  booktitle=ICCV,
  pages={8858--8867},
  year={2019}
}

@inproceedings{deng2009imagenet,
  title={Imagenet: A large-scale hierarchical image database},
  author={Deng, Jia and Dong, Wei and Socher, Richard and Li, Li-Jia and Li, Kai and Fei-Fei, Li},
  booktitle=CVPR,
  pages={248--255},
  year={2009}
}

@inproceedings{isola2017image,
  title={Image-to-image translation with conditional adversarial networks},
  author={Isola, Phillip and Zhu, Jun-Yan and Zhou, Tinghui and Efros, Alexei A},
  booktitle=CVPR,
  pages={1125--1134},
  year={2017}
}

@inproceedings{liu2015deep,
  title={Deep learning face attributes in the wild},
  author={Liu, Ziwei and Luo, Ping and Wang, Xiaogang and Tang, Xiaoou},
  booktitle=ICCV,
  pages={3730--3738},
  year={2015}
}

@article{karras2017progressive,
  title={Progressive growing of gans for improved quality, stability, and variation},
  author={Karras, Tero and Aila, Timo and Laine, Samuli and Lehtinen, Jaakko},
  journal={arXiv preprint arXiv:1710.10196},
  year={2017}
}

@article{zhou2017places,
  title={Places: A 10 million image database for scene recognition},
  author={Zhou, Bolei and Lapedriza, Agata and Khosla, Aditya and Oliva, Aude and Torralba, Antonio},
  journal=PAMI,
  volume={40},
  number={6},
  pages={1452--1464},
  year={2017},
}

@article{doersch2012what,
  title={What Makes Paris Look like Paris?},
  author={Carl Doersch and Saurabh Singh and Abhinav Gupta and Josef Sivic and Alexei A. Efros},
  journal=SIGGRAPH,
  volume={31},
  number = {4},
  pages={101:1--101:9},
  year={2012},
}

@article{kingma2014adam,
  title={Adam: A method for stochastic optimization},
  author={Kingma, Diederik P and Ba, Jimmy},
  journal={arXiv preprint arXiv:1412.6980},
  year={2014}
}

@article{wang2004image,
  title={Image quality assessment: from error visibility to structural similarity},
  author={Wang, Zhou and Bovik, Alan C and Sheikh, Hamid R and Simoncelli, Eero P},
  journal=TIP,
  volume={13},
  number={4},
  pages={600--612},
  year={2004},
}

@inproceedings{heusel2017gans,
  title={Gans trained by a two time-scale update rule converge to a local nash equilibrium},
  author={Heusel, Martin and Ramsauer, Hubert and Unterthiner, Thomas and Nessler, Bernhard and Hochreiter, Sepp},
  booktitle=NIPS,
  pages={6626--6637},
  year={2017}
}

@inproceedings{szegedy2016rethinking,
  title={Rethinking the inception architecture for computer vision},
  author={Szegedy, Christian and Vanhoucke, Vincent and Ioffe, Sergey and Shlens, Jon and Wojna, Zbigniew},
  booktitle=CVPR,
  pages={2818--2826},
  year={2016}
}

@InProceedings{shiftnet2018,
author="Yan, Zhaoyi
and Li, Xiaoming
and Li, Mu
and Zuo, Wangmeng
and Shan, Shiguang",
title="Shift-Net: Image Inpainting via Deep Feature Rearrangement",
booktitle=ECCV,
year="2018",
pages="3--19"
}

@article{yu2015multi,
  title={Multi-scale context aggregation by dilated convolutions},
  author={Yu, Fisher and Koltun, Vladlen},
  journal={arXiv preprint arXiv:1511.07122},
  year={2015}
}

@article{canny1986computational,
  title={A computational approach to edge detection},
  author={Canny, John},
  journal=PAMI,
  number={6},
  pages={679--698},
  year={1986},
}

@inproceedings{wan2020bringing,
  title={Bringing Old Photos Back to Life},
  author={Wan, Ziyu and Zhang, Bo and Chen, Dongdong and Zhang, Pan and Chen, Dong and Liao, Jing and Wen, Fang},
  booktitle=CVPR,
  pages={2747--2757},
  year={2020}
}

@article{danon2019unsupervised,
  title={Unsupervised natural image patch learning},
  author={Danon, Dov and Averbuch-Elor, Hadar and Fried, Ohad and Cohen-Or, Daniel},
  journal={Computational Visual Media},
  volume={5},
  number={3},
  pages={229--237},
  year={2019},
  publisher={Springer}
}

@article{hu2020Jittor,
  title={Jittor: A Noval Deep Learning Framework with Unified Graph Execution and Meta Operators},
  author={Hu, Shi-Min and Liang, Dun and Yang, Guo-Ye and Yang, Guo-Wei and Zhou, Wen-Yang},
  journal={Science China-Information Sciences},
  year={2020},
  note={to appear. \href{https://github.com/Jittor/Jittor}{https://github.com/Jittor/Jittor}.},
}

@unpublished{quick2018,
  title={Quick Draw Irregular Mask Dataset},
  author={Iskakov, Karim},
  year=2018,
  note ={\href{https://github.com/karfly/qd-imd}{https://github.com/karfly/qd-imd}. Accessed 5 Match 2020.},
}

@unpublished{officail2019,
  title={Official implement for Learning Pyramid-Context Encoder Network for High-Quality Image Inpainting},
  author={Zeng, Yanhong},
  year=2019,
  note ={\href{https://github.com/researchmm/PEN-Net-for-Inpainting}{https://github.com/researchmm/PEN-Net-for-Inpainting}. Accessed 5 Match 2020.},
}

@unpublished{deepfill2019,
  title={DeepFill v1/v2 with Contextual Attention and Gated Convolution},
  author={Yu, Jiahui},
  year=2019,
  note ={\href{https://github.com/JiahuiYu/generative\_inpainting}{https://github.com/JiahuiYu/generative\_inpainting}. Accessed 5 Match 2020.},
}

@unpublished{unofficial2019,
  title={Unofficial implementation of "Image Inpainting for Irregular Holes Using Partial Convolutions"},
  author={Gruber, Mathias},
  year=2019,
  note ={\href{https://github.com/MathiasGruber/PConv-Keras}{https://github.com/MathiasGruber/PConv-Keras}. Accessed 5 Match 2020.},
}

%-------------------------------------------------------------------------

\end{document}